\pdfoutput=1
\documentclass[11pt]{article}

\usepackage{acl}

\usepackage{times}
\usepackage{latexsym}

\usepackage[T1]{fontenc}
\usepackage[utf8]{inputenc}

\usepackage{microtype}

\usepackage[T1]{fontenc}    % use 8-bit T1 fonts
\usepackage{hyperref}       % hyperlinks
\usepackage{url}            % simple URL typesetting
\usepackage{booktabs}       % professional-quality tables
\usepackage{amsfonts}       % blackboard math symbols
\usepackage{nicefrac}       % compact symbols for 1/2, etc.
\usepackage{microtype}      % microtypography
\usepackage{xcolor}         % colors
\usepackage{graphicx}
\usepackage{float}
\usepackage{ulem}
\usepackage{algorithm,algcompatible,amsmath}
\algnewcommand\INPUT{\item[\textbf{Input:}]}%
\algnewcommand\OUTPUT{\item[\textbf{Output:}]}%
\title{APAM: Adaptive Pre-training and Adaptive Meta Learning in Language Model for Noisy Labels and Long-tailed Learning}

\author{Sunyi Chi \\
  \texttt{sunyi.chi@uth.tmc.edu} \\\And
  Bo Dong\\
  \texttt{dongbd@amazon.com} \\\And
  Yiming Xu\\
  \texttt{ymxu@amazon.com} \\
  \AND
  Zhenyu Shi\\
  \texttt{szhenyu@amazon.com} \\\And
  Zheng Du\\
  \texttt{zhengdu@amazon.com} \\}

\begin{document}
\maketitle
\begin{abstract}
Practical natural language processing (NLP) tasks are commonly long-tailed with noisy labels. Those problems challenge the generalization and robustness of complex models such as Deep Neural Networks (DNNs). Some commonly used resampling techniques, such as oversampling or undersampling, could easily lead to overfitting. It is growing popular to learn the data weights leveraging a small amount of metadata. Besides, recent studies have shown the advantages of self-supervised pre-training, particularly to the under-represented data. In this work, we propose a general framework to handle the problem of both long-tail and noisy labels. The model is adapted to the domain of problems in a contrastive learning manner. The re-weighting module is a feed-forward network that learns explicit weighting functions and adapts weights according to metadata. The framework further adapts weights of terms in the loss function through a combination of the polynomial expansion of cross-entropy loss and focal loss. Our extensive experiments show that the proposed framework consistently outperforms baseline methods. Lastly, our sensitive analysis emphasizes the capability of the proposed framework to handle the long-tailed problem and mitigate the negative impact of noisy labels.
\end{abstract}

\section{Introduction}
Deep Neural Networks (DNNs) have become the default modeling choice for complex problem with large-scale labeled data. They have been remarkably successful in supervised learning across a variety of domains such as natural language processing and computer vision. Their success relies on the availability of a large amount of labelled data with high quality. In practice, it is usually expensive to acquire clean labels at scale. It either requires multiple blind passes and adjudicators decision or needs quality assurance by auditor. Both are labor intensive and time-consuming. Recent progress on fine tuning~\cite{bert, finetune2}, domain adaptation~\cite{adda, davqa, dann, osda} and few-shot learning~\cite{fewshot} alleviate the demand for large volume labeled data. Deep learning models remain dependent on accurate labeled data, in spite of those progress. 

Another factor that compromises model generalization is data distribution shift. Data shift can occur from various sources. Long-tailed problem is a common example. For example, in order to train a model to classify shopping items, it is difficult to obtain sufficient images for rare products~\cite{amznprod}. It is also a challenge to train a dialog model that is exposed to sufficient less frequent topics or user intents. When training a model on an imbalanced dataset, model training becomes biased towards the majority classes. With higher number of examples available to learn from, the model learns to perform well on the majority classes but due to the lack of enough examples the model fails to learn meaningful patterns that could aid it in learning the minority classes. The model performance on those tail classes bottlenecks the applications of deep neural networks in practice and thus it is critical to improve on such cases. 

A number of studies have proposed approaches to mitigate noisy label or long-tailed class problem. To alleviate impact of noisy label, sample selection~\cite{ss2}, label correction~\cite{losscorrect1, losscorrect2}, and noise-aware losses~\cite{noiseloss1, noiseloss2} have been studied. Dataset resampling such as SMOTE~\cite{smote} is a popular method by selecting a proportion of data to train a network or by learning a weight for each example. The weights are optimized by minimizing the training loss. It is applied in multiple well-known algorithms such as AdaBoost~\cite{AdaBoost}, self-paced learning~\cite{selfpacedlearn}, and algorithm that emphasizes high variance samples~\cite{ss2, highvar2}. To address problem of long-tailed distributions, some studies have modified sampling algorithm to ensure all classes are represented equally~\cite{repequal, smote}. Other popular approaches include adjusting loss function~\cite{llloss} biased to minor class, and post-hoc correction~\cite{posthoccor}.

However, those methods have contradicting assumptions. On the one hand, we assign higher weight to clean labelled data in order to mitigate noisy label problem. On the other hand, algorithms for long-tailed problems emphasize minority classes that more likely have higher training loss. Therefore, those methods can not handle the problem of concurrent noisy label and long-tailed classes. In order to handle noisy label and long tailed problem simultaneously, some works~\cite{mwn, learnreweight} propose a meta-learning paradigm that follows a more natural assumption that the best example weighting should minimize the loss of clean data. Those methods learn instance weights from a small clean dataset and show promising results. Nevertheless, we argue that the meta-learning paradigm is not general enough, and remains expensive due to the requirement of a balanced meta validation dataset.

In our work, we propose the Adaptive Pre-training and Adaptive Meta Learning method (APAM), a general framework to handle the concurrent problems of noisy label and long-tailed classes together. It naturally subsumes the aforementioned meta-learning paradigm~\cite{mwn, learnreweight} as a special case. Our method does not require a balanced meta data to guide reweighting. This fact reduces the amount of clean data needed in meta learning and thus is more feasible and less expensive. 

Furthermore, we introduce a stage of domain adaptive pre-training~\cite{domainadapt} through contrastive learning~\cite{simcse} to improve model robustness. On the one hand, the proposed adaptive procedure give lower weights on noisy samples; on the other hand, APAM give higher weights to simultaneously handle the noise and long-tailed problems.

We further evaluate APAM on two datasets in which it outperforms other methods~\cite{mwn, focal, cbloss, bert, simcse}. In addition, we conduct comprehensive ablation study and sensitivity analysis. Across those experiments, we observe APAM consistently outperforms other methods.

To summarize, the main contributions of our work are listed as follows:

\begin{itemize}
\item To cope with the concurrent problems of long-tail and noisy label in text classification, we propose a general two-stage deep learning framework including domain adaptive pre-training stage and supervised fine-tuning through adaptive re-weighting stage. It outperforms the state-of-the-art methods in evaluation datasets.
\item We demonstrate that the proposed adaptive weighting method does not require balanced meta data and thus alleviate the dependence to large amount of meta data in long-tailed problem.
\item Our experiment results illustrate that domain adaptive contrastive learning consistently leads to improvement of performance in APAM framework.
\item Through a holistic ablation study and sensitivity analysis, we demonstrate the contribution of each components of the proposed method and the effectiveness of this method to long-tailed problem with noisy label.
\end{itemize}

\section{Related Work}

\subsection{Noisy label learning}

Recent success of deep learning is highly dependent on a massive data with high quality label. However, accurate labels are expensive and labor-intensive to obtain. Some crowdsourcing platforms, such as Amazon Mechanical Turk, have been widely used to reduce labeling cost. Those solutions, however, often results in corrupted labels~\cite{mturk1, mturk2}. Song et al.~\cite{noisysurvey} summarized that the noisy ratio of labels commonly ranges from 8.0\% to 38.5\%~\cite{range1, range2, range3} in real-world. In the presence of corrupted labels, deep learning models tend to overfit the noisy labels. Zhang et al.~\cite{fitnoisy} proved in experiment that deep neural networks can easily fit a training dataset with a variety of noisy ratio, which results in poor generalization of model.

Many studies have explored noisy label learning that can be categorized into four buckets including regularization, design loss, sampling method, and modify model architecture to resist to noisy data. 

Specifically, regularization methods are widely adopted including dropout~\cite{dropout}, weight decay~\cite{weightdecay}, and batch normalization~\cite{batchnorm}, as well as implicit regularization such as label smoothing~\cite{lsmooth1, lsmooth2}, mixup~\cite{mixup} and adversarial training~\cite{adverexample}. 

Many studies have proposed robust loss~\cite{robloss1, robloss2, robloss3} or loss adjustment~\cite{adjloss1, range1, losscorrect1, wu2023bottrinet, losscorrect2}, methods to alleviate noisy label problem. Meta learning~\cite{learnreweight, mwn, metalearning, cml} is a type of automated weighting adjustment or label correction method through learning to learn how to reweight or correct labels. Sampling methods~\cite{range1,ss1,sp3}, focus on selecting true-labeled examples from a noisy training dataset. Moreover, model architectures~\cite{modelart1, modelart2} that resists to noisy label have been widely applied such as noise adaptation layer.

\subsection{Long-tailed learning}

The real-world applications are commonly long-tailed problems where some classes are associated with very few data points~\cite{cbloss, llloss, lslt}.  Long-tailed problem is different with class-imbalanced problem in two ways. First, the minority class can have a large number of samples in spite of relatively smaller size compared to majority class. Second, long-tailed problems have a large number of classes and the tail-class samples are often very scarce. In contrast, the number of classes can be very small in class imbalanced problem. Therefore, long-tailed problem can be regarded as a more specific and challenging class imbalance problem~\cite{deeplongtailsurvey}.

A massive deep long-tailed learning methods have been proposed in recent years~\cite{cbloss, lslt} to address long-tailed problem. Those methods can be categorized to three types~\cite{deeplongtailsurvey} including information augmentation, class re-balancing and model improvement. Information augmentation based methods aim at improving long-tailed learning by introducing additional information into model training. Popular methods include transfer learning~\cite{ltmtt, sdtl} and data augmentation~\cite{suridadl}. Class Re-balancing methods consist of re-sampling~\cite{smote, posthoccor, blossnlp} and loss adjusting methods~\cite{polyloss, focal, csbcid}. Lastly, improving model is also proved useful in many long-tailed problems. Popular methods include ensemble learning~\cite{ensemblesoft} and representation learning~\cite{clbhybrid, wu2023fineehr, speed, bigdatasecurity}.

\subsection{Domain adaptive pre-training}
Pretrained language models from a wide variety of corpus have reformed NLP since recent years. Devlin et al.,~\cite{bert} proposed self-supervised learning (SSL), an unsupervised learning method that lets model learn data representations by solving auxiliary tasks without labels. Those pre-training tasks can improve model generalization and prevent the model from being overfitted to a limited number of class labels~\cite{cml}. 

Moreover, many studies~\cite{domainadapt, selfregtc} have demonstrated the benefit to adapt a pretrained model to the domain of target tasks via a second phase of in-domain pretraining. Gururangan et al.~\cite{domainadapt} present a study of eight classification tasks across four different domains of publications such as biomedical science, computer science, news, and show that the domain-adaptive pretraining (DAPT) leads to performance improvement. Zhang at al.~\cite{mglongtaillearner} have focused on making pre-trained language models suitable for long-tailed problem.

Contrastive learning~\cite{contlearn1st} has been widely applied in pre-training tasks. The key idea is to learn representation by pulling semantically close neighbors together and pushing apart non-neighbors. Researchers~\cite{simcse, clear} have proposed to learn sentence embeddings through contrastive learning. This recent work shows that contrastive pre-training on unsupervised tasks at scale leads to effective representations of text and code~\cite{tcembedding}.

\section{Methodology}
Inspired by recent studies in contrastive learning and meta learning, we propose APAM, a general two-stage framework to handle long tailed NLP problem in noisy data setting. We assume the problem contains only a small set of clean data and large set of noisy labeled data. 

APAM first leverages a domain adaptive pre-training stage through contrastive learning in a self-supervised manner. The SSL task is to learn semantically close feature representations of pair sentences through random mask of dropout layer where the positive pair is generated from the same source sentence. 

\begin{figure*}
  \centering
  \includegraphics[scale=0.69]{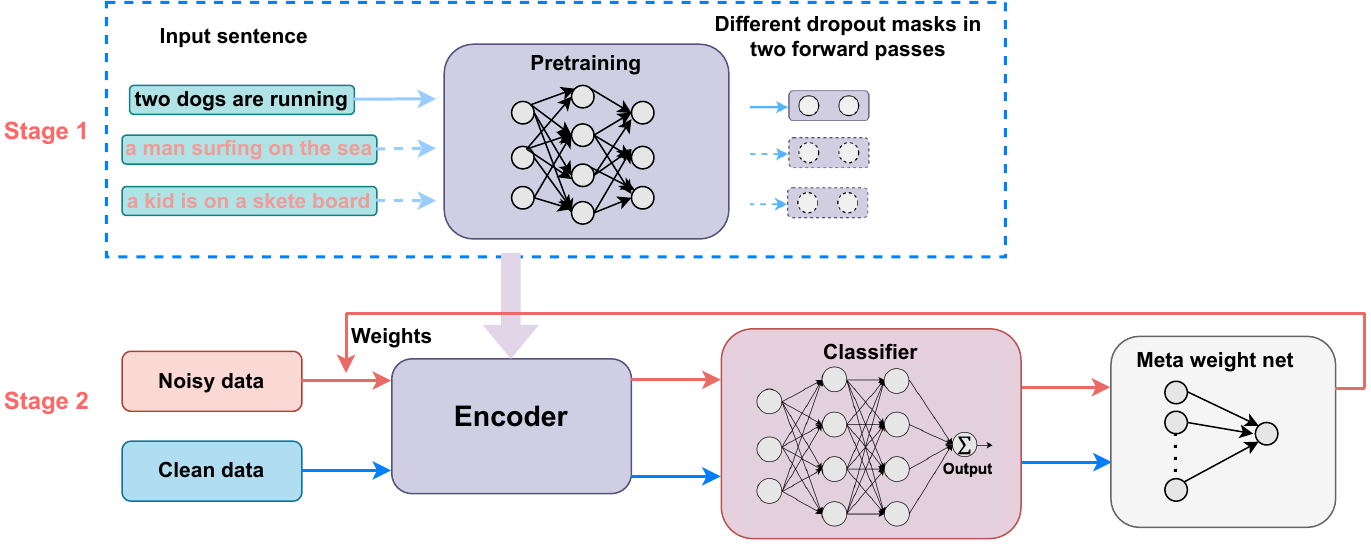}
  \caption{An overview of APAM framework, where $\theta$ denotes parameters for meta model, and $W$ denotes parameters for the main model.}
  \label{framework}
\end{figure*}

In the second stage, we propose adaptive weighting strategies that leverage a small amount of clean data to advise the adaptation of encoder parameters. Iteratively, the noisy train sample will gain the weights from forward pass and leverage those weights for the consecutive iteration. The process can mitigate the problem of long-tailness and noisy labels. The proposed framework is shown in the Figure~\ref{framework}. Below we describe the two aforementioned main components in detail.

%Draw.io

\subsection{Domain adaptive pre-training}
The first stage of our framework pre-trains the base model with self-supervised learning, thereby discarding the noisy and unbalanced labels. We utilize a simple contrastive learning of sentence embeddings (simCSE)~\cite{simcse} to learn domain specific representations in the pre-training stage. We fine tune the base encoder to the specific domain of task through adaptive training on the entire set of unlabeled sentences $\{x_i\}$. Here, we use  identical sentence pair $(x_i, x_i^+)$ as positive pairs, where $x_i^+ = x_i$, through independently sampled dropout masks. Specifically, we feed the same input to the encoder $F(x, z)$ twice and get two embeddings with different dropout masks $z$ and $z'$, where $z$ is denoted random mask for dropout. We denote feature representations of sentence identical pair $(x_i, x_i^+)$ as $(s_i, s_i^+)$, where $s_i = F(x, z)$ and $s_i^+ = F(x^+, z')$. Other different in-batch sentences are considered as negative pairs. The contrastive loss is shown in the Equation~\ref{eq:contras_loss}.

\begin{equation}
\label{eq:contras_loss}
l_i = -log\frac{e^{sim(s_i, s_i^+)/\tau}}{\sum^N_{j=1}{e^{sim(s_i, s_j^+)/\tau}}}
\end{equation}

Let $N$ denote the number of sentences in a mini-batch. $\tau$ is a temperature parameter that is fixed to $\tau= 0.05$ following the previous study~\cite{simcse}; $sim(s_1, s_2)$ is the cosine similarity $\frac{s_1^Ts_2}{||s_1||||s_2||}$. In this stage, we obtain a domain adaptive pre-trained encoder by large amount of unlabeled text in the specific domain.

\subsection{Adaptive meta learning}
In the second stage, we conduct adaptive meta learning leveraging a small amount of clean set to fine-tune the base model iteratively. 
Inspired by Polyloss \cite{polyloss}, we use an adaptive polynomial
term on cross-entropy loss shown in Equation~\ref{eq:polyloss}
\begin{equation}
\label{eq:polyloss}
l(W)=-log(P_c) + \epsilon(1-P_c)
\end{equation}

where $\epsilon$ is the weight of polynomial term which is 1 in our experiments, $P_c$ stands for predicted probability of the target class, and $W$ denotes set of parameters including model parameters and adaptive term in loss function. 

% whole APAM algorithm
\begin{algorithm}
    
    \caption{Adaptive meta learning algorithm}
    \begin{algorithmic}[1]
    \REQUIRE Training data $D_t$, meta data $D_m$.
    \INPUT Batch size n,m, max iterations $T$.
    \OUTPUT $W_T$
    \STATE \textbf{Initialization} $\theta_0$, $W_0$
    \FOR{$t$ = 0 , ... , $T$-1}
      \STATE $\{x,y\} \gets$ SampleMiniBatch$(D_t, n)$
      \STATE $\{x^{meta},y^{meta}\} \gets$SampleMiniBatch$(D_m, m)$
      \STATE $\hat{y} \gets $Forward$(x,y,W_t)$
      \STATE $\nabla W_t \gets $Backward$(W_t(\theta))$
      \STATE $\hat{W_t}(\theta) \gets W_t-\alpha\nabla W_t$
      \STATE $\hat{y^{meta}} \gets $Forward$(x^{meta},y^{meta},\hat{W_t}(\theta))$
      \STATE $\nabla \theta \gets $Backward$(\theta)$
      \STATE update $\hat{\theta}$
      \STATE $\nabla W_t \gets $Backward$(W_t(\hat{\theta}))$
      \STATE $W_{t+1} \gets$ OptimizerStep($W_t,\nabla W_t$)
    \ENDFOR
  \end{algorithmic}
  \label{algo}
\end{algorithm}

Inspired by some existing works ~\cite{learnreweight, mwn}, we use a MLP network with only one hidden layer containing 100 nodes inputting loss $L^{train}(p)$ to automatically learn instance weights. This model requires two nested loops of optimization, as shown in the Algorithm~\ref{algo}. Specifically, in each iteration of training, we update the training model parameter $W$ by moving the current $W$ along the descent direction of the objective loss with initialized meta model parameter $\theta$, where $\alpha$ is the step size. After receiving the updated $W(\theta)$, the parameters $\theta$ of the meta model then are updated by moving the current parameter along the objective gradient calculated on the meta-data. Then, the updated $\hat{\theta}$ is employed to update the parameter $W(\hat{\theta})$  through propagation.

\section{Experiments}
To evaluate the performance of APAM, we conduct experiments on multiple classification tasks in two real-world datasets. We also compare APAM with the state-of-the-art approaches of learning with long-tail and noisy labels.

\subsection{Dataset}
\paragraph{Amazon annotation} This Amazon annotation dataset is sampled from Alexa live traffic and is an enterprise-scale spoken language data of dialog interaction between human and conversational AI agent. Please note that we have processed the data so that users are not identifiable. The dataset includes Automatic Speech Recognition (ASR) text for utterance, %labeled goal category 
annotated by human annotator. Annotators listen the audios and select labels from pre-defined taxonomy of categories. There are 30 plus different classes for this task. 
%We illustrate a few examples in Table~\ref{datasets2}.
%Utterances from some classes are more prevalent than others such as music and shopping while we also demonstrate examples of minor class such as Gallery. 
More rigorously, we define imbalance factor of a dataset as the number of training samples in the largest class divided by the smallest class. The imbalance factor of this dataset is 110. 
%This data contains real world noisy ratio and imbalance rate. 

In addition, the data contains 6\% noisy labels where we collect noisy label from an annotation pipeline where only one individual annotator annotates the domain of the utterance. In contrast, we attain clean label from another data pipeline where three annotators provide the label based on the majority vote or an additional adjudicator if annotators can not reach an agreement. 

% \begin{table*}[!htbp]
% \centering
%  \caption{Dataset}
%  \label{datasets}
% \begin{tabular}{ccccc}
% \hline
% Dataset    & Alexa Goal &&Amazon Review& \\ \hline
% \# classes&30&&29 \\
% Imbalance&110&&51.3\\
% Noisy&$\sim$6\%&&10\%, 20\%, 30\%, 40\%, 50\%, 60\%\\
% \hline
% %\multicolumn{3}{l}{\small *THIS IS A NICE FOOTNOTE.} \\
% \end{tabular}
% \end{table*}

\begin{table*}[!htbp]
\centering
 \caption{Example classes and texts of Amazon Review dataset}
 \label{datasets2}
\begin{tabular}{ccc}
\hline
    Example classes&Example texts \\ \hline
%Alexa Goal&Physical shopping& where is bread \\
%& Celebrity voices& can i talk with kim\\
%&Gallery& do you know when is edvard munch's gallery \\
%&Navigation& navigate to burger king \\

\hline
Books&loved the story \\
Clothing Shoes and Jewelry&comfortable and gorgeous\\
Patio, Lawn and Garden&looks beautiful and fits perfectly in our backyard\\
Music instruments&very well made and purchased as a gift \\
\hline
%\multicolumn{3}{l}{\small *THIS IS A NICE FOOTNOTE.} \\
\end{tabular}
\end{table*}
\begin{table*}[!htbp]
 \caption{Model performance on Amazon Annotation dataset}
 \label{main}
 \centering
\begin{tabular}{lccccc}
\hline
\multicolumn{1}{c}{(\%)}                                    & Accuracy       & \multicolumn{1}{l}{$\Delta$Accuracy}          & Precision      & Recall         & F1-score       \\ \hline
{\color[HTML]{000000} BERT + cross entropy loss} & 83.33          & {\color[HTML]{000000} + 0}    & 83.13          & 83.33          & 82.79          \\
BERT   + focal loss                                         & 83.66          & {\color[HTML]{000000} + 0.33} & 83.38          & 83.66          & 82.90           \\
BERT   + class-balanced loss                                & 83.97          & {\color[HTML]{000000} + 0.64} & 83.45          & 83.97          & 83.05          \\
BERT   + MWN                                                & 84.07          & {\color[HTML]{000000} + 0.74} & 83.93          & 84.07          & 83.44          \\
BERT   + simCSE                                             & 84.08          & {\color[HTML]{000000} + 0.75} & 83.90           & 84.08          & 83.25          \\
{\color[HTML]{000000} BERT + AML}                           & {\color[HTML]{000000} 84.22} & {\color[HTML]{000000} + 0.89} & {\color[HTML]{000000} 83.90} & {\color[HTML]{000000} 84.22} & {\color[HTML]{000000} 83.50} \\
{\color[HTML]{000000} BERT   + DAPT-SSL}                    & {\color[HTML]{000000} 84.60} & {\color[HTML]{000000} + 1.27} & {\color[HTML]{000000} 84.39} & {\color[HTML]{000000} 84.60} & {\color[HTML]{000000} 83.78} \\
BERT   + APAM                                               & \textbf{85.38} & {\color[HTML]{000000} + 2.05} & \textbf{84.77} & \textbf{85.38} & \textbf{84.56} \\ \hline
\end{tabular}
\end{table*}

\paragraph{Amazon review} In addition to the enterprise-level Amazon annotation dataset, We also evaluate APAM on dataset Amazon review\footnote{https://jmcauley.ucsd.edu/data/amazon/}, which is one of the largest product review datasets. It is widely used with imbalanced classification benchmark methods. We compare APAM with multiple baseline approaches such as BERT~\cite{bert}, MWNet~\cite{mwn}, etc., and conduct sensitivity analysis by varying the noisy rates. This dataset contains Amazon product reviews and metadata across 19 years, including 142.8 million reviews for 29 product categories such as Books, Electronics, Beauty, etc. Reviews include textual comments, ratings, and votes, product metadata such as descriptions, category information, price, and brand. The largest class contains over 200 times more reviews than the minority class. We randomly sampled subset of the data in which all users and items have at least 5 reviews. The imbalance factor is 51.3. 
%Information of this dataset is summarized in 
Table~\ref{datasets2} shows 2 example head and 2 example tail categories of Amazon Review dataset with their corresponding texts.

% the  of the proposed APAM on a widely used benchmark datasets Amazon reviews. 
% The input feature of Amazon review is product reviews from Amazon.

% \begin{table*}
% \centering
%  \caption{Dataset}
%  \label{datasets}
% \begin{tabular}{cccccc}
% \hline
% Dataset    & \# classes & Imbalance&Noisy&Example Category&Example text \\ \hline
% Alexa Goal& 30 &  110&\sim6\%& Music&play hip hop\\
% &&&&Smart home&is A/C opened\\
% Amazon Review&29&51.3&10\%-60\%&Clothing Shoes and Jewelry& comfortable and gorgeous\\
% &&&&Books&loved the story \\
% \hline
% %\multicolumn{3}{l}{\small *THIS IS A NICE FOOTNOTE.} \\
% \end{tabular}
% \end{table*}

\subsection{Experimental setting}
\paragraph{Synthesize noise ratio} In Amazon reviews data, we sampled 1\% of the entire training set as the clean set. The noisy sets are generated by corrupting the labels of remaining data points based on Uniform label noise. We generate the noisy ratio $\rho = (10\%, 20\%, 30\%, 40\%, 50\%, 60\%)$. For a dataset with $C$ classes, a clean example with true label y is randomly corrupted to all possible classes with probability $\rho/C$. Noted that the label also has probability of $\rho/C$ to stay truth, hence the corrupted label might also happen to be the original label.

\paragraph{Model architectures and training}
We use uncased Bert-base as encoder backbone throughout the entire experiment. Two fully connected layers are put on top of the pre-trained BERT-base with $256$ and $128$ hidden nodes. Training batch size is set to $128$ with initial learning rate $1e^{-5}$. We implement all models and experiments in PyTorch. All models are trained with the same number of epochs.

\paragraph{Baseline methods}
We evaluate APAM against BERT and meta weight net (MWN) as baseline methods. MWN is a meta-learning paradigm that follows a natural assumption that the best example weighting should minimize the loss of clean data, which learns instance weights from a small clean dataset and works remarkably well in practice. Moreover, we conduct a thorough ablation study to prove the positive impact of domain adaptive pre-training and adaptive meta learning.

%\begin{table*}[!htbp]
% \caption{Model performance on Amazon Annotation dataset}
% \label{main}
%\begin{tabular}{lccccc}
%\hline
%\multicolumn{1}{c}{(\%)}                                    & Accuracy       & \multicolumn{1}{l}{$\Delta$Accuracy}          & Precision      & Recall         & F1-score       \\ \hline
%{\color[HTML]{000000} BERT + cross entropy loss} & 83.33          & {\color[HTML]{000000} + 0}    & 83.13          & 83.33          & 82.79          \\
%BERT   + focal loss                                         & 83.66          & {\color[HTML]{000000} + 0.33} & 83.38          & 83.66          & 82.90           \\
%BERT   + class-balanced loss                                & 83.97          & {\color[HTML]{000000} + 0.64} & 83.45          & 83.97          & 83.05          \\
%BERT   + MWN                                                & 84.07          & {\color[HTML]{000000} + 0.74} & 83.93          & 84.07          & 83.44          \\
%BERT   + simCSE                                             & 84.08          & {\color[HTML]{000000} + 0.75} & 83.90           & 84.08          & 83.25          \\
%BERT   + APAM                                               & \textbf{85.38} & {\color[HTML]{000000} + 2.05} & \textbf{84.77} & \textbf{85.38} & \textbf{84.56} \\ \hline
%\end{tabular}
%\end{table*}

\subsection{Main results}
We used metrics: accuracy, precision, recall and F1-score for performance evaluation. Table~\ref{main} presents the accuracy, precision, recall, F1-score of APAM  with pre-trained BERT-base as its main classifier, comparing with other state-of-art methods on Amazon Annotation dataset with real world imbalanced and noisy labels. The baseline model BERT-base with cross entropy loss achieves 83.33\% accuracy, 83.13\% precision, 83.33\% recall and 82.79\% F1-score. APAM achieves 85.38\% accuracy, 84.77\% precision, 85.38\% recall and 84.56\% F1-score. Compared to baseline model, APAM improves 2.05\% accuracy taking advantage of adaptive instance weight learning from a small clean set in the noisy unbalanced Amazon Annotation dataset. 

\begin{table*}[!htbp]
  \caption{Comparison of APAM against human annotators.}
  \label{class-table}
  \centering
\begin{tabular}{clcccccc}
\hline
      &                     & \multicolumn{3}{c}{APAM}                               & \multicolumn{3}{c}{Human}    \\ \hline
      &                     & Precision        & Recall            & F1              & Precision & Recall  & F1     \\ \hline
      & Physical Shopping   & 91.32          & 98.34           & 94.70          & 93.54   & 99.24 & 96.30 \\
Major & Celebrity Voices    & 90.88          & 82.52           & 86.50          & 97.36   & 99.01 & 98.18 \\
Class & Command and Control & 88.73          & 93.91           & 91.25          & 94.59   & 96.19 & 95.39 \\
      & Music               & 93.27          & 96.52           & 94.87          & 95.87   & 97.60 & 96.73 \\ \hline
      & Podcasts            & \textbf{90.24} & \textbf{97.37}  & \textbf{93.67} & 89.47   & 89.47 & 89.47 \\
Minor & Recipes             & 76.67          & \textbf{95.83}  & 85.19          & 100.00  & 91.67 & 95.65 \\
Class & Gallery             & \textbf{94.44} & \textbf{94.44}  & \textbf{94.44} & 94.44   & 94.44 & 94.44 \\
      & Navigation          & 85.71          & \textbf{100.00} & \textbf{92.31} & 100.00  & 83.33 & 90.91 \\ \hline
\end{tabular}
\end{table*}

\begin{table*}[!htbp]
  \caption{Sensitivity analysis results on Amazon review dataset}
  \label{sample-table2}
  \centering
  \begin{tabular}{lccccccc}
\hline
\multicolumn{1}{c}{(Noise ratio \%)}                                    & 10                                   & \multicolumn{1}{l}{20}                & 30                                    & 40                                    & 50                                   & \multicolumn{1}{l}{60}                & \multicolumn{1}{l}{Average}           \\ \hline
{\color[HTML]{000000} BERT + cross entropy loss} & {\color[HTML]{000000} 61.37}         & {\color[HTML]{000000} 59.57}          & {\color[HTML]{000000} 57.09}          & {\color[HTML]{000000} 55.04}          & {\color[HTML]{000000} 52.77}         & {\color[HTML]{000000} 22.58}          & {\color[HTML]{000000} 51.40}          \\
{\color[HTML]{000000} BERT + MWN}                           & {\color[HTML]{000000} 61.53}         & {\color[HTML]{000000} 61.13}          & {\color[HTML]{000000} 59.91}          & {\color[HTML]{000000} 57.60}           & {\color[HTML]{000000} 55.70}          & {\color[HTML]{000000} 53.18}          & {\color[HTML]{000000} 58.18}          \\
BERT   + APAM                                               & {\color[HTML]{000000} \textbf{65.70}} & {\color[HTML]{000000} \textbf{65.15}} & {\color[HTML]{000000} \textbf{64.15}} & {\color[HTML]{000000} \textbf{63.18}} & {\color[HTML]{000000} \textbf{62.70}} & {\color[HTML]{000000} \textbf{61.63}} & {\color[HTML]{000000} \textbf{63.75}} \\ \hline
\end{tabular}
\end{table*}

Furthermore, Table~\ref{class-table} shows APAM prediction performance against human annotator's annotation, where we observe APAM achieves a comparable performance against human annotator. Noticeably, APAM outperforms human labeling in some tail classes such as Podcasts and Navigation categories. Those observations implies the potential of label automation and particularly emphasis the under-represented classes.

%In real world long-tailed dataset, the limited ground truth data are also long-tailed. Existing noisy learning methods \cite{learnreweight, mwn} require a balanced clean dataset as meta data to learn weights. \textcolor{red}{[What does this mean? ]Although they only require a small amount of balanced meta data, it still requires a large sample size of ground truth data due to the limited sample sizes in tail classes of real world long-tailed dataset. }Therefore, we also conduct experiment to explore the effect of imbalanced meta data on APAM. The experiment results show that the APAM using the adaptive meta learning method is robust on unbalanced meta data so that it does not require a balanced meta dataset. Applying to Alexa Annotation dataset, APAM shows 85.38\% accuracy either using balanced or unbalanced meta data. 

\subsection{Ablation study}
To evaluate the contribution of each components of the APAM framework, we conduct this holistic ablation study to understand what is the actually effect of each component and quantify the benefit of the APAM framework. Results are  in Table~\ref{main}. The domain adaptive pre-training in self-supervised learning (DAPT-SSL), using unlabeled text additionally pre-training ahead of the baseline model, improves 1.27\% accuracy over baseline. The adaptive meta learning, taking advantage of a small clean set to weight each instance, achieved 84.22\% accuracy. The APAM takes advantage of both DAPT-SSL and AML and achieves 2.05\% accuracy improvement over baseline model on Amazon Annotation dataset, which demonstrates the effectiveness of each components of APAM under real world long-tailed with noisy labels dataset. 

%\begin{table*}[!htbp]
%  \caption{Ablation study results of APAM components}
%  \label{sample-table1}
%  \centering
%  \begin{tabular}{lccccc}
%\hline
%\multicolumn{1}{c}{(\%)}                                    & Accuracy                     & \multicolumn{1}{l}{$\Delta$Accuracy}          & Precision                    & Recall                       & F1-score                     \\ \hline
%{\color[HTML]{000000} BERT + cross entropy loss} & 83.33                        & {\color[HTML]{000000} + 0}    & 83.13                        & 83.33                        & 82.79                        \\
%{\color[HTML]{000000} BERT + AML}                           & {\color[HTML]{000000} 84.22} & {\color[HTML]{000000} + 0.89} & {\color[HTML]{000000} 83.90} & {\color[HTML]{000000} 84.22} & {\color[HTML]{000000} 83.50} \\
%{\color[HTML]{000000} BERT   + DAPT-SSL}                    & {\color[HTML]{000000} 84.60} & {\color[HTML]{000000} + 1.27} & {\color[HTML]{000000} 84.39} & {\color[HTML]{000000} 84.60} & {\color[HTML]{000000} 83.78} \\
%BERT   + APAM                                               & \textbf{85.38}               & {\color[HTML]{000000} + 2.05} & \textbf{84.77}               & \textbf{85.38}               & \textbf{84.56}               \\ \hline
%\end{tabular}
%\end{table*}

\subsection{Sensitivity analysis}

Table~\ref{sample-table2} presents the sensitivity analysis results using benchmark dataset Amazon reviews under UNIF noise levels (10\%, 20\%, 30\%, 40\%, 50\% and 60\%) and imbalance factor 51. It’s clear that, since baseline method does not handle the noisy and imbalanced labels, its performance decreases significantly when the noise level goes
up and becomes extremely worse as the noise level turns to 60\%; while re-weighting based methods (MWN and APAM) show robustness against severe label noises for the long-tailed data. This is consistent with results of image data
reported in~\cite{mwn} where re-weighting
was shown to perform well either in varying noise rate or in real world imbalanced rate. Moreover, we observe that APAM is more effective in doing this than previous re-weighting methods for noisy and imbalanced text data. The more severe the noisy issue is, the more advantage APAM has over baseline and other re-weighting methods. 

\section{Conclusion}
In this paper, we address the problem of learning with noisy
labels and long-tailed text data from an adaptive weight learning and meta-learning perspective. We propose a novel framework for adaptively pre-training unlabeled domain specific text and extracting sample weights to guarantee robust deep learning in the presence of training data bias. 
Our empirical results show that the
proposed framework APAM outperforms other state-of-the-art methods for long-tailed and noisy learning areas on both real world datasets and synthetic datasets. The proposed adaptive framework APAM is robust against different level of noise and imbalanceness. In addition, it is compatible with other baseline architectures.
%besides can be generalized to use on other baseline architectures.

%\section*{Acknowledgements}

%\clearpage
%\bibliographystyle{abbrv}
\bibliography{egbib}

\bibliographystyle{acl_natbib}

\end{document}